\documentclass{article}
\usepackage{ijcai24}

\usepackage{times}
\usepackage{soul}
\usepackage{url}
\usepackage[hidelinks]{hyperref}
\usepackage[utf8]{inputenc}
\usepackage[small]{caption}
\usepackage{graphicx}
\usepackage{amsmath}
\usepackage{amsthm}
\usepackage{booktabs}
\usepackage{algorithm}
\usepackage{algorithmic}
\usepackage[switch]{lineno}

\usepackage{comment}
\usepackage{multirow}
\usepackage{multicol}
\usepackage{bm}
\usepackage{ amssymb, amsfonts}
\usepackage{subcaption}
\usepackage{array} 
\usepackage{verbatim}
\title{Multi-Scale Dilated Convolution Network for Long-Term Time Series Forecasting}

\author{
Feifei Li$^{1,2}$
\and
Suhan Guo$^{1,3}$\and
Feng Han$^{1,2}$\And
Jian Zhao$^{4},^*$\\
Furao Shen$^{1,3},^*$
\affiliations
$^1$National Key Laboratory for Novel Software Technology,Nanjing University,China\\
$^2$Department of Computer Science and Technology,Nanjing University,China\\
$^3$School of Artificial Intelligence,Nanjing University,China\\
$^4$School of Electronic Science and Engineering,Nanjing University,China\\
\emails
\{feifeili, shguo,fenghan\}@smail.nju.edu.cn,
\{jianzhao,frshen\}@nju.edu.cn
}

\begin{document}
\maketitle
\begin{abstract}
Accurate forecasting of long-term time series has important applications for decision making and planning. 
However, it remains challenging to capture the long-term dependencies in time series data.
To better extract long-term dependencies, We propose \underline{M}ulti-\underline{S}cale \underline{D}ilated \underline{C}onvolution \underline{N}etwork (\textbf{MSDCN}), a method that utilizes a  shallow dilated convolution architecture to capture the period and trend characteristics of long time series. 
We design different convolution blocks with exponentially growing dilations and varying kernel sizes to sample time series data at different scales. 
Furthermore, we utilize traditional autoregressive model to capture the linear relationships within the data. 
To validate the effectiveness of the proposed approach, we conduct experiments on eight challenging long-term time series forecasting benchmark datasets.
The experimental results show that our approach outperforms the prior state-of-the-art approaches and shows significant inference speed improvements compared to several strong baseline methods. 
\end{abstract}

\section{Introduction}

Time series forecasting is widely used in real life, including power consumption prediction ~\cite{pang2018hierarchical}, economy forecasting ~\cite{qin2017dual}, weather prediction ~\cite{karevan2020transductive},  traffic prediction ~\cite{wu2020connecting}, and so on. 
Among these prediction demands, it is often necessary to forecast over a longer time window to make timely decisions or give early warnings. 
Through the analysis and modeling of historical data, long-term time series forecasting enables us to gain insights and forecast future trends, cycles, and potential influencing factors. 
For instance, by forecasting power consumption, the power system can be efficiently managed to ensure smooth operations and meet the demands of both production and daily life. 
At the same time, the speed of forecasting is critical. 
For example, traffic flow forecasting needs to display the real-time road occupancy for the next few minutes, so as to optimize the allocation of road resources and alleviate congestion. 
Therefore, timely and accurate time series forecasting plays a crucial role in practical applications.

\begin{figure}[tbp]
\centerline{\includegraphics[scale=0.5]{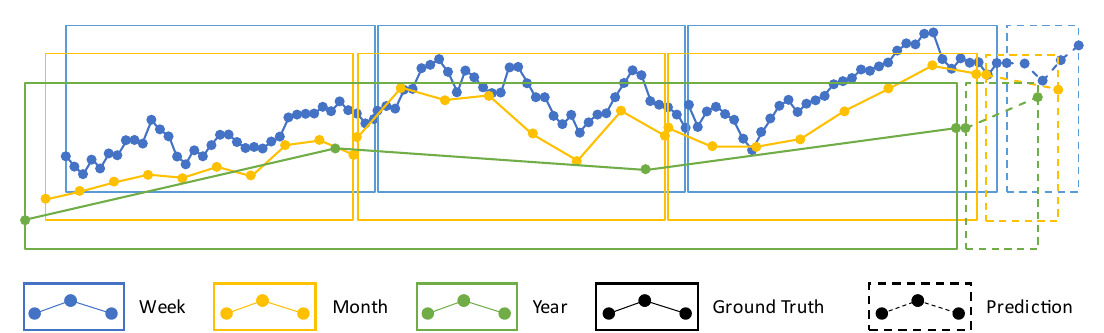}}
%\centerline{\includegraphics[scale=0.9]{./fig/network/motivation.png}}
% 考虑一种比较理想的情况，一条曲线的周期性和趋势性都比较明显，我们人眼观察的时候会每隔N个点观察一次，总结规律
% 一条时间序列可以看作是周趋势、月趋势、年趋势、随机噪声等成分的线性或非线性叠加。
\caption{A time series can be considered as the linear or nonlinear superposition of components such as weekly trends, monthly trends, and yearly trends.}
%\caption{The motivation of our methods. When people observe at intervals of 24 data points visually, they can analyze and summary the patterns. Based on the observed patterns, they can deduce the likely performance of the next 24th data points. The red points represent the data we have observed, while the green points represent our reasonable predictions.}
\label{fig:motivation}
\end{figure}

Times series can be regarded as the linear or nonlinear superposition of components such as weekly trends, monthly trends, yearly trends and random noise, as depicted in Figure \ref{fig:motivation}. The key to time series modeling is to efficiently extract these components and model subsequent tasks based on the extracted components. 
Some methods ~\cite{zengAreTransformersEffective2022,wuAutoformerDecompositionTransformers2021} based on time series decomposition apply operations like moving averages and pooling to decompose time series, partially ignoring small but crucial information. 
Moreover, the extraction operations are inflexible, making it challenging to capture multi-scale information.

Traditional time series forecasting methods ~\cite{arima1970time,taylorForecastingScale2017} mainly adopt functions fitting on sequence data. 
However, these traditional methods have difficulties in dealing with big data and complex temporal patterns. 
In addition, methods based on recurrent neural network ~\cite{qin2017dual,shihTemporalPatternAttention2019,salinasDeepARProbabilisticForecasting2020} are prone to the problem of vanishing gradient, which makes it difficult to effectively capture long-term information. 

For long-term time series forecasting, there are three main types of methods using neural networks: Transformer-based methods, MLP-based methods and CNN-based methods.
Transformer ~\cite{vaswaniAttentionAllYou2017} is a commonly-used model for sequence modeling. 
Nevertheless, the permutation invariance property inherent in its self-attention mechanism poses challenges in capturing the temporal characteristics of time series data effectively ~\cite{wen2023transformers}. This limitation hinders the extraction of precise temporal representations from time series data. Moreover, it is worth noting that Transformer-based models exhibit high time and space complexity, which can be a concern in resource-constrained scenarios. 
Many methods ~\cite{liEnhancingLocalityBreaking2019,kitaevReformerEfficientTransformer2019,liuNonstationaryTransformersExploring2022,liuPyraformerLowComplexityPyramidal2022,zeng2022muformer} focus on improving the redundant self-attention mechanism and point-to-point connections in the transformer. 
MLP-based methods ~\cite{oreshkinNBEATSNeuralBasis2020,challuNHiTSNeuralHierarchical2022} first decouple data into trend and remainder using moving average operations and then forecast them separately. However, the extracted patterns are fixed and relatively single.
For example,  in DLinear ~\cite{zengAreTransformersEffective2022}, the kernel size of the moving average operation is fixed, and it can only extract one scale of feature representation.
CNN-based methods ~\cite{laiModelingLongShortTerm2018,wuTimesNetTemporal2DVariation2023}
 use ordinary convolution operations to extract local dependencies, which can only extract short-term local dependencies.
Some methods ~\cite{wangMICNMultiscaleLocal2023} employ causal convolution ~\cite{vanwavenet} to extract long-term global dependencies, but these methods lack the flexibility to extract multi-scale time information effectively.

In summary, current methods still face two major challenges when dealing with long-term time series forecasting. 
Firstly, long-term time series data often exhibit complex periodic and trend changes, making it difficult for existing methods to capture both the short-term local dependency and the long-term global dependency. 
Secondly, existing methods have high time complexity and space complexity, which may pose limitations in terms of computing and storage resources in real-world application scenarios. 

% 贡献点：
To solve the two challenges mentioned above, we propose a multi-scale dilated convolution shallow network (\textbf{MSDCN}). 
Dilated convolution downsamples the time series at different scales.
To capture a broader range of scales, we expand the receptive field of the convolution network by using different kernel sizes and exponentially changing the dilation size. 
In addition, our model incorporates the traditional autoregressive module to directly model the linear relationship of data and facilitate complex modeling. 
The contributions of our paper are summarized as follows:
\begin{itemize}
    \item We propose a novel approach that utilize  the intrinsic characteristics of dilated convolution to expand the receptive field for time series data. 
    \item We introduce a multi-scale feature fusion  shallow structure, which consists of two levels of multi-scale features. These features involve an exponentially growing number of dilations and different kernel sizes. 
    \item  We leverage traditional autoregressive models to capture linear dependencies inherent in the data, effectively streamlining the non-linear modeling process. 
    \item We conduct  experiments to demonstrate the effectiveness of each module in our model. Our method achieves state-of-the-art performance. Furthermore, our model exhibits superior speed in both the training and inference phases.
\end{itemize}

\section{Related Works} \label{2}
%Long-term series forecasting always focus on the modeling of long-term dependency information and relationships among multiple variables recently. 
\subsection{Long-term Dependency Information}
Transformer-based methods focus on reducing quadratic time complexity and quadratic memory usage of the canonical self-attention mechanism when modeling the long sequence time series.
%Informer
Informer ~\cite{zhouInformerEfficientTransformer2021} proposes ProbSparse self-attention mechanism to calculate the most similar queries to participate in the attention computation through KL divergence, hence reducing unnecessary calculations.
However, these methods are still restricted by the permutation invariance characteristics of the Vanilla Transformer ~\cite{vaswaniAttentionAllYou2017}. The numerical values in time series data often lack semantics,  and the crucial aspect of modeling is to capture the temporal changes between consecutive sequences of points.
% Autoformer
To preserve time information, Autoformer ~\cite{wuAutoformerDecompositionTransformers2021} proposes  Auto-Correlation mechanism to replace self-attention mechanism and discover sequence-level dependencies through Fast Fourier transform. 

CNN-based models  use convolutional neural networks ~\cite{lecun1998gradient} to extract both local and global patterns. 
% LSTNet
LSTNet ~\cite{laiModelingLongShortTerm2018} employs CNNs to capture short-term local dependency patterns and uses RNNs ~\cite{elman1990finding,memory2010long,chung2014empirical} to capture long-term patterns. However, the design of the skip length $p$ in the model can only extract a single scale of temporal pattern and becomes a bottleneck for modeling.
MICN ~\cite{wangMICNMultiscaleLocal2023} uses mutil-scale isometric convolution to extract local and global features. The isometric convolution is similar to TCN  ~\cite{EmpiricalEvaluationGeneric2018baia}. 
However, the modeling maybe restricted by causal convolution for time series forecasting since the input does not contain future sequences. 
To overcome the limitations of TCN, SCINet ~\cite{liuSCINetTimeSeries2021} proposes downsample convolution and interaction architecture to extract rich features from multiple resolutions. 
TimesNet ~\cite{wuTimesNetTemporal2DVariation2023} uses a parameter-efficient inception block to extract complex temporal variations. 
However, those models usually have high space complexity and time complexity.
\begin{figure*}[hbtp]
\centerline{\includegraphics[width=\textwidth]{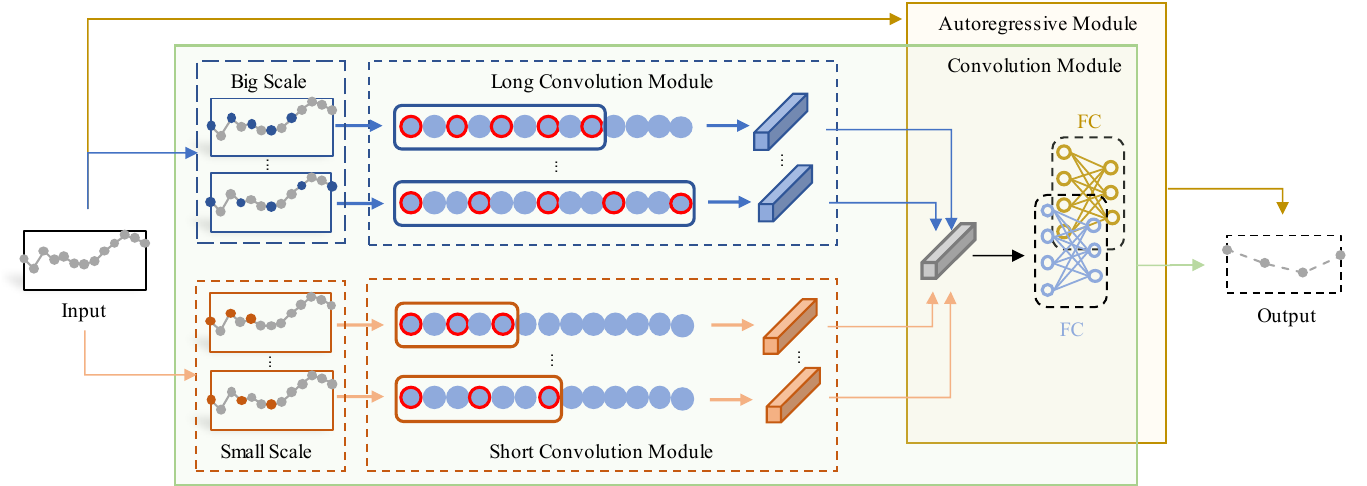}}
\caption{The overall architecture of MSDCN. The green box represents convolution
 module, and the yellow box represents autoregressive module.}
\label{fig:network}
\end{figure*}

\subsection{The Relationship Between Multiple Variables}
Recently, many methods have studied how to model the relationship between multiple variables.
%LightTS
LightTS ~\cite{zhangLessMoreFast2022} uses a linear layer to do channel projection, and conduct experiments to verify a linear layer is sufficient to model the variable relationship. 
%Crossformer
Crossformer ~\cite{zhangCrossformerTransformerUtilizing2023} proposes a Two-stage Attention Layer, which calculates attention in the time dimension and the space dimension separately. 
%DLinear and PatchTST
Due to the phenomenon of distribution drift in the time series, DLinear ~\cite{zengAreTransformersEffective2022} and TimesNet ~\cite{wuTimesNetTemporal2DVariation2023} independently model multiple variables, and also achieve good performance. 
Recently, ~\cite{CapacityRobustnessTradeoff2023han} conducts experiments and provides theoretical evidence that independent prediction can alleviate the issue of distribution drift. Inspired by those methods, we also adopt the channel independency strategy in our model to achieve more robust predictions.

\section{Method} \label{3}
\subsection{Problem Formulation}
Long-term time series forecasting can be formally defined as follows: the input is 
\begin{equation}
    \mathbf{X}=\{X_1^t, X_2^t, \cdots, X_C^t\}_{t=1}^T \in \mathbb{R}^{T\times C}, 
\end{equation}
where $T$ is the lookback window size, $C$ is the number of variables, $X_i^t$ represents the value of the $i$th variable at time $t$. The output is 
\begin{equation}
    \mathbf{Y}=\{X_1^t, X_2^t, \cdots, X_C^t\}_{t=T+1}^{T+L}  \in \mathbb{R}^{L\times C}, 
\end{equation}
where $L$ is the prediction window size, We use previous $T$ steps to predict future $L$ steps, where $L> T$ in our research problem. Our target is to learn a mapping function $\mathbf{f}$ to predict future time series, which is 
\begin{equation}
    \mathbf{Y=f(X)}.
\end{equation}

\subsection{Overview}
Originating from image processing, dilated convolution ~\cite{yu2017dilated} aims to solve the problem of information loss caused by traditional convolution or pooling during downsampling process.
By introducing gaps (dilations) between convolution kernels, dilated convolution enlarges the receptive field of the convolution operation, allowing it to better capture pixel information from more distant locations. 
This property enables the network to effectively acquire more details during downsampling, thus alleviating the problem of information loss.

This paper proposes a method that utilizes dilated convolution to extract multi-scale information from time series through downsampling. 
The dilation rates in different convolution kernels grow exponentially, providing a more flexible method to capture information from various scales of sequence components.
%写清每个模块是怎么连接的，以及简要地描述他们的作用
%我们的模型由非常简单的结构构成，图1展示了整个结构。多变量输入序列被分别送到长期卷积模块和短期卷积模块进行特征提取。长期卷积模块和短期卷积模块的区别在于他们的感受野，也即kernel size是不同的。每个卷积模块由多个不同的卷积block组成。在卷积模块提取到序列特征后，使用不同的权重对其进行特征融合，得到融合后的特征表示。融合后的特征，经过一层前馈神经网络，与自回归得到的部分相加得到最终的结果。

We introduce a multi-scale dilated convolution structure, as shown in Figure \ref{fig:network}.
Convolution module downsamples the input data $\mathbf{X}$ at various scales.
%The input $\mathbf{X}$ is passed through both a long-term convolutional module and a short-term convolutional module for feature extraction. 
Long and short convolutional module comprises multiple blocks, which enable the generation of multi-scale feature representations.
We then employ different weights for feature fusion. 
The fused features are further processed by a feed forward neural network layer. 
The final prediction is obtained as the sum of the output from the feed forward layer and the autoregressive component. 
In addition, following the NLinear model ~\cite{zengAreTransformersEffective2022}, we add normalization operations before and after the model, which is not shown in Figure \ref{fig:network}.

\subsection{Convolution Module}
% 讲明变化的参数和可调的参数。
%卷积模块由N个不同的1维卷积块并列组成。N是可调的参数。每个一维卷积块由1维卷积+BatchNorm+ReLU激活函数组成。其中，1维卷积是一个分通道的空洞卷积。正如论文PatchTST中提到的，把每个变量独立地预测反而会得到更好的效果。因此，我们采用了组数为变量数的分通道卷积。同时，为了模拟自相关系数的计算来进行特征提取并且引入非线性，我们使用了空洞卷积。假设空洞数为K，则代表卷积核每隔K个点观察一次。不同的空洞大小代表了不同尺度的特征提取。N个并列的1维卷积的空洞数以2的倍数增长，即$dilations={2^0+1, 2^1+1, ...2^N+1}$。
We use two distinct convolutional modules with the same underlying structure but different convolution kernel sizes to downsample the input data at different scales. 
The long-term convolutional module employs a larger kernel size, representing a larger receptive field, while the short-term module utilizes a smaller kernel size, representing a smaller receptive field. 
Each convolutional module consists of multiple different 1D convolution blocks arranged in parallel.
%as depicted in Figure \ref{fig2}. 
Specifically, assuming the long convolution module comprises $n$ convolution blocks, the short convolution module has $m$ convolution blocks. These blocks respectively produce
\begin{equation}
    h_i=\mathrm{ConvBlock}_i(\mathbf{X}) \quad i=1, \cdots, n, n+1\cdots, n+m, 
\end{equation}
where $h_i \in \mathbb{R}^{C \times T}$ is the output of the $i$th ConvBlock. More specifically, each convolution block is a sequence structure composed of dilated convolution, batch normalization and relu activation function, that is
\begin{equation}
\begin{aligned}
    u_i&=\mathrm{DilatedConv1d}( \mathrm{Padding} ( \mathbf{X} ) ), \\
    h_i&=\mathrm{ReLU}( \mathrm{BatchNorm1d}( u_i ) ).
\end{aligned}
\end{equation}

The one-dimensional convolution is a depthwise convolution, where the number of input channels and output channels are the number of variables in the datasets.
At the same time, we use dilated convolution to extract long-term time information.
The dilation size denotes that the convolution kernel observes at a fixed interval.
Different dilation sizes represent different scales of feature extraction. 
To gradually expand the receptive field, the number of dilation sizes in the parallel one-dimensional convolutions increases by a factor of 2, that is, 
\begin{equation}
dilation \  factors=\{2^0+1, 2^1+1, ..., 2^n+1\}.
\end{equation}
%我们采用了两种不同的卷积模块，一种是长期卷积模块，即1维卷积的卷积核大小为25。一种是短期卷积模块，即卷积核大小为7。对于卷积核的大小，我们并没有做特殊的设计，相对来讲，25代表了一种更长期的提取，7代表了一种较短期的提取。然后我们对不同的ConvBlock训练学习了不同的权重，从而进行特征表示的融合。

Then for different convolution blocks, the  weights $W\in \mathbb{R}^{C \times (n+m)}$ are learnt to get fusion feature representations $M$ from different scales, as shown in Equation \ref{eq:M},
\begin{equation}
\begin{aligned}
    H&=\left[h_1, h_2, \cdots, h_{n+m}\right], \\
    M&=\sum_i^{n+m} H_i \odot W^{\prime}_i, 
\end{aligned}
\label{eq:M}
\end{equation}
where $H \in \mathbb{R}^{C \times T \times (n+m)}$ is the stack of $h_i$, $W^{\prime}$ is a repeat of $W$ along $T$ dimension, $\odot$ denotes the element-wise product, and $M\in \mathbb{R}^{C\times T}$ is the fusion feature representation. Then the fusion feature representation goes through a layer of feed forward neural network that produces:
\begin{equation}
    \hat{Y}_c=W_1M^T+b_1, 
    \label{yc}
\end{equation}
where $W_1 \in \mathbb{R}^{L\times T} $ denotes the linear weights, $b_1 \in \mathbb{R}^{L}$ is a bias parameters, $\hat{Y}_c$ is the first part of final prediction.

\subsection{Autoregressive Module}
% 对比方法LSTNet, NLinear, LightTS, 产生反作用，我们的可以起到正向作用
% 时间序列数据中往往有一些简单的重复模式，因此为了让神经网络更容易学到重复模式，我们添加了自回归预测的模块，类似于ResNet的思想。LighTS和LSTNet模型都加入了这一模块，但是不同的是，他们的模型甚至要比单一的Linear模型表现更差，起到了负作用。而我们的模型起到了正向的作用，即能够使得预测更加准确。
To enhance the learning of repetitive patterns commonly observed in time series data, our model incorporates an classical autoregressive prediction module, which follows a similar concept to ResNet ~\cite{resnet}. Autoregressive model is a regression process concerning the variable itself, capable of capturing underlying linear dynamic relationships in time series.
\begin{equation}
    Y_t=\sum_{i=1}^p \phi_iY_{i-1}+\epsilon_t
\end{equation}
where $\phi_i$ are coefficients and $\epsilon_t$ is white noise error. In the MSDCN architecture, we use $W$ as the coefficients of the autoregressive model.
\begin{equation}
    \hat{Y}_h=W_2\mathbf{X}+b_2, 
    \label{yh}
\end{equation}
where $W_2 \in \mathbb{R}^{L \times T}$ is the coefficient of the autoregressive operation, $b_2 \in \mathbb{R}^{L}$ is a bias parameter, $\hat{Y}_h$ denotes the second part of final prediction. Compared to the LightTS ~\cite{zhangLessMoreFast2022} and LSTNet ~\cite{laiModelingLongShortTerm2018} models which also incorporate this module, our model shows a positive effect and generates more accurate predictions. Detailed analysis is given in Section \ref{exp}.

Finally, the final prediction of our model is 
\begin{equation}
    \hat{Y}=\hat{Y}_c+\hat{Y}_h.
\end{equation}
where $\hat{Y}_c$ is given in Equation \ref{yc}, $\hat{Y}_h$ is given in Equation \ref{yh}.

To reduce the influence of abnormal outliers, We choose Huber Loss \cite{huber1964robust} as the loss function in the training process, as follows:
\begin{equation}
loss=\begin{cases}
0.5\ (y_i-\hat y_i)^2, &if \  \left| y_i-\hat y_i\right| < \delta \\
\delta *(\left|y_i-\hat y_i\right|-0.5*\delta),  &\text{otherwise}, 
\end{cases}
\label{eq3}
\end{equation}
where $\delta$ is a positive and adjustable threshold. 

\section{Experiments}\label{exp}
\subsection{Setup}

% datasets, split, seq_len
\textbf{Dataset} We conduct experiments on eight benchmark datasets for time series forecasting, including 
(1) ETT datasets ~\cite{zhouInformerEfficientTransformer2021} are collected from electricity transformers, including oil temperature and load.
(2) Electricity dataset ~\cite{ECL} contains electricity consumption from 321 clients. 
(3) Illness dataset ~\cite{Illness}  contains influenza-like illness patients data. 
(4) Weather dataset ~\cite{Weather}  comprises 21 weather indicators  from the Weather Station of the Max Planck Biogeochemistry Institute. 
(5) Traffic dataset ~\cite{Traffic} contains the road occupancy rates from 862 sensors. 

The characteristics of each dataset are shown in Table \ref{tab1}. However, we do not conduct experiments on Exchange-rate dataset, due to the reasons metioned in PatchTST ~\cite{nieTimeSeriesWorth2022}. 
To ensure the fairness of our experiments, we strictly follow the standard protocol for dataset splitting. 
Specifically, we divide all datasets into training, validation, and test sets in chronological order. 
For the ETT dataset, we use a split ratio of 6:2:2, while for other datasets, we adopt a ratio of 7:1:2.  
For the illness dataset, the model's input length is 36 and its output length can be 24, 36, 48 or 60. 
For the other datasets, the model's input length is fixed at 96, and its output length can be 96, 192, 336 or 720.
To evaluate our model performance, we adopt two commonly used metrics: mean squared error (MSE) and mean absolute error (MAE). MSE and MAE are computed as:
\begin{equation}
    MSE=\frac{1}{N}\sum_{j=1}^N ( \hat{y}_{j}-y_{j})^2, 
    \label{eq1}
\end{equation}
\begin{equation}
    MAE=\frac{1}{N} \sum_{j=1}^N  \left| \hat{y}_{j}-y_{j}\right |, 
    \label{eq2}
\end{equation}
where $N$ is the number of variables, $\hat{y}$ is the prediction and $y$ is the ground truth.

\begin{table}[tbp]
  \centering
  \caption{Experimental Datasets}
    \resizebox{0.8\linewidth}{!}{
    \begin{tabular}{c|c|c|c}
    \toprule
    Datasets & Variables & Frequency & Observations \\
    \midrule
    ETTm1 & 7   & 15 Minutes & 69,680 \\
    ETTm2 & 7   & 15 Minutes & 69,680 \\
    ETTh1 & 7   & 1 hour & 17,420 \\
    ETTh2 & 7   & 1 hour & 17,420 \\
    \midrule
    Illness & 7   & 1 week & 966 \\
    Weather & 21  & 10 Minutes & 52,696 \\
    Electricity & 321 & 1 hour & 26,304 \\
    Traffic & 862 & 1 hour & 17,544 \\
    \bottomrule
    \end{tabular}%
    }

  \label{tab1}%
\end{table}%
\begin{comment}
\textbf{Metrics} To evaluate our model performance, we adopt two commonly used metrics: mean squared error (MSE) and mean absolute error (MAE). MSE and MAE are computed as:
\begin{equation}
    MSE=\frac{1}{N}\sum_{j=1}^N ( \hat{y}_{j}-y_{j})^2, 
    \label{eq1}
\end{equation}
\begin{equation}
    MAE=\frac{1}{N} \sum_{j=1}^N  \left| \hat{y}_{j}-y_{j}\right |, 
    \label{eq2}
\end{equation}
where $N$ is the number of variables, $\hat{y}$ is the prediction and $y$ is the ground truth.
\end{comment}

\begin{table*}[htbp]
  \centering
  \caption{Multivariate long-term forecasting results.}
  \resizebox{\textwidth}{!}{
  
    \begin{tabular}{c|c|cc|cc|cc|cc|cc|cc|cc|cc}
    \midrule
    \multicolumn{2}{c|}{Models}& \multicolumn{2}{c|}{MSDCN}& \multicolumn{2}{c|}{TimesNet}& \multicolumn{2}{c|}{NLinear}& \multicolumn{2}{c|}{DLinear}& \multicolumn{2}{c|}{LightTS}& \multicolumn{2}{c|}{ETSformer}& \multicolumn{2}{c|}{Autoformer}& \multicolumn{2}{c}{Informer} \\
    \midrule
    \multicolumn{2}{c|}{Metric}& MSE& MAE& MSE& MAE& MSE& MAE& MSE& MAE& MSE& MAE& MSE& MAE& MSE& MAE& MSE& MAE \\
    \midrule
\multirow{4}[2]{*}{\rotatebox{90}{ETTm2}}& 96& \textbf{0.174}& \textbf{0.256}& 0.187& 0.267& \underline{0.182}& \underline{0.265}& 0.193& 0.292& 0.209& 0.308& 0.189& 0.280& 0.255& 0.339& 0.365& 0.453  \\
  & 192& \textbf{0.239}& \textbf{0.298}& 0.249& 0.309& \underline{0.246}& \underline{0.304}& 0.284& 0.362& 0.311& 0.382& 0.253& 0.319& 0.281& 0.340& 0.533& 0.563  \\
  & 336& \textbf{0.296}& \textbf{0.332}& 0.321& 0.351& \underline{0.306}& \underline{0.341}& 0.369& 0.427& 0.442& 0.466& 0.314& 0.357& 0.339& 0.372& 1.363& 0.887  \\
  & 720& \textbf{0.395}& \textbf{0.392}& 0.408& 0.403& \underline{0.408}& \underline{0.398}& 0.554& 0.522& 0.675& 0.587& 0.414& 0.413& 0.433& 0.432& 3.379& 1.338  \\ %\cmidrule{2-18}
 % & Avg& \textbf{0.276}& \textbf{0.320}& 0.291& 0.333& \underline{0.286}& \underline{0.327}& 0.350& 0.401& 0.409& 0.436& 0.293& 0.342& 0.327& 0.371& 1.410& 0.810  \\
    \midrule
    \multirow{4}[2]{*}{\rotatebox{90}{ETTh1}}& 96& \textbf{0.379}& \textbf{0.390}& \underline{0.384}& 0.402& 0.393& 0.400& 0.386& \underline{0.400}& 0.424& 0.432& 0.494& 0.479& 0.449& 0.459& 0.865& 0.713  \\
  & 192& \textbf{0.428}& \textbf{0.417}& \underline{0.436}& \underline{0.429}& 0.449& 0.433& 0.437& 0.432& 0.475& 0.462& 0.538& 0.504& 0.500& 0.482& 1.008& 0.792  \\
  & 336& \textbf{0.465}& \textbf{0.436}& 0.491& 0.469& 0.485& \underline{0.449}& 0.481& 0.459& 0.518& 0.488& 0.574& 0.521& 0.521& 0.496& 1.107& 0.809  \\
  & 720& \textbf{0.468}& \textbf{0.453}& 0.521& 0.500& \underline{0.471}& \underline{0.462}& 0.519& 0.516& 0.547& 0.533& 0.562& 0.535& 0.514& 0.512& 1.181& 0.865  \\ %\cmidrule{2-18}
  %& Avg& \textbf{0.435}& \textbf{0.424}& 0.458& 0.450& \underline{0.449}& \underline{0.436}& 0.456& 0.452& 0.491& 0.479& 0.542& 0.510& 0.496& 0.487& 1.040& 0.795  \\
    \midrule
    \multirow{4}[2]{*}{\rotatebox{90}{ECL}}& 96& \underline{0.175}& \textbf{0.265}& \textbf{0.168}& \underline{0.272}& 0.198& 0.275& 0.197& 0.282& 0.207& 0.307& 0.187& 0.304& 0.201& 0.317& 0.274& 0.368  \\
  & 192& \textbf{0.183}& \textbf{0.271}& \underline{0.184}& 0.289& 0.198& \underline{0.278}& 0.196& 0.285& 0.213& 0.316& 0.199& 0.315& 0.222& 0.334& 0.296& 0.386  \\
  & 336& \underline{0.199}& \textbf{0.287}& \textbf{0.198}& 0.300& 0.212& \underline{0.293}& 0.209& 0.301& 0.230& 0.333& 0.212& 0.329& 0.231& 0.338& 0.300& 0.394  \\
  & 720& \underline{0.238}& \textbf{0.320}& \textbf{0.220}& \textbf{0.320}& 0.254& 0.326& 0.245& 0.333& 0.265& 0.360& 0.233& 0.345& 0.254& 0.361& 0.373& 0.439  \\ %\cmidrule{2-18}
  %& Avg& \underline{0.199}& \textbf{0.286}& \textbf{0.192}& 0.295& 0.216& \underline{0.293}& 0.212& 0.300& 0.229& 0.329& 0.208& 0.323& 0.227& 0.338& 0.311& 0.397  \\
    \midrule
    \multirow{4}[2]{*}{\rotatebox{90}{Traffic}}& 96& 0.619& \underline{0.366}& \textbf{0.593}& \textbf{0.321}& 0.647& 0.388& 0.650& 0.396& \underline{0.615}& 0.391& 0.607& 0.392& 0.613& 0.388& 0.719& 0.391  \\
  & 192& \textbf{0.577}& \underline{0.342}& 0.617& \textbf{0.336}& 0.600& 0.364& \underline{0.598}& 0.370& 0.601& 0.382& 0.621& 0.399& 0.616& 0.382& 0.696& 0.379  \\
  & 336& \textbf{0.591}& 0.348& 0.629& \textbf{0.336}& 0.607& 0.367& \underline{0.605}& 0.373& 0.613& 0.386& 0.622& 0.396& 0.622& \underline{0.337}& 0.777& 0.420  \\
  & 720& \textbf{0.630}& \underline{0.365}& 0.640& \textbf{0.350}& 0.645& 0.387& 0.645& 0.394& 0.658& 0.407& \underline{0.632}& 0.396& 0.660& 0.408& 0.864& 0.472  \\ %\cmidrule{2-18}
  %& Avg& \textbf{0.604}& \underline{0.355}& \underline{0.620}& \textbf{0.336}& 0.625& 0.376& 0.625& 0.383& 0.622& 0.392& 0.621& 0.396& 0.628& 0.379& 0.764& 0.416  \\
    \midrule
    \multirow{4}[2]{*}{\rotatebox{90}{Weather}}& 96& \textbf{0.169}& \textbf{0.217}& \underline{0.172}& \underline{0.220}& 0.202& 0.240& 0.196& 0.255& 0.182& 0.242& 0.197& 0.281& 0.266& 0.336& 0.300& 0.384  \\
  & 192& \textbf{0.215}& \textbf{0.259}& \underline{0.219}& \underline{0.261}& 0.248& 0.277& 0.237& 0.296& 0.227& 0.287& 0.237& 0.312& 0.307& 0.367& 0.598& 0.544  \\
  & 336& \textbf{0.269}& \textbf{0.299}& \underline{0.280}& \underline{0.306}& 0.300& 0.313& 0.283& 0.335& 0.282& 0.334& 0.298& 0.353& 0.359& 0.395& 0.578& 0.523  \\
  & 720& \underline{0.350}& \textbf{0.352}& 0.365& \underline{0.359}& 0.373& 0.360& \textbf{0.345}& 0.381& 0.352& 0.386& 0.352& 0.388& 0.419& 0.428& 1.059& 0.741  \\ %\cmidrule{2-18}
  %& Avg& \textbf{0.251}& \textbf{0.282}& \underline{0.259}& \underline{0.287}& 0.281& 0.298& 0.265& 0.317& 0.261& 0.312& 0.271& 0.334& 0.338& 0.382& 0.634& 0.548  \\
    \midrule
    \multirow{4}[2]{*}{\rotatebox{90}{ILI}}& 24& \textbf{2.222}& \underline{0.936}& \underline{2.317}& \textbf{0.934}& 2.662& 1.054& 2.398& 1.040& 8.313& 2.144& 2.527& 1.020& 3.483& 1.287& 5.764& 1.677  \\
  & 36& \underline{2.192}& \textbf{0.915}& \textbf{1.972}&\underline{0.920}& 2.487& 1.040& 2.646& 1.088& 6.631& 1.902& 2.615& 1.007& 3.103& 1.148& 4.755& 1.467  \\
  & 48& \textbf{2.164}& \textbf{0.938}& \underline{2.238}& \underline{0.940}& 2.406& 1.024& 2.614& 1.086& 7.299& 1.982& 2.359& 0.972& 2.669& 1.085& 4.763& 1.469  \\
  & 60& \underline{2.287}& \underline{0.946}& \textbf{2.027}& \textbf{0.928}& 2.475& 1.037& 2.804& 1.146& 7.283& 1.985& 2.487& 1.016& 2.770& 1.125& 5.264& 1.564  \\ %\cmidrule{2-18}
 % & Avg& \underline{2.216}& \underline{0.934}& \textbf{2.139}& \textbf{0.931}& 2.507& 1.039& 2.616& 1.090& 7.382& 2.003& 2.497& 1.004& 3.006& 1.161& 5.137& 1.544  \\ 
    \bottomrule
    \end{tabular}%
   % }
}
  \label{tab:main1}%
\end{table*}%
\textbf{Baselines} We compare our method with  the following baselines: 
\begin{itemize}
    \item Transformer-based models: Informer ~\cite{zhouInformerEfficientTransformer2021}, Autoformer ~\cite{wuAutoformerDecompositionTransformers2021}, ETSformer ~\cite{ETSformerExponentialSmoothing2022woo}.
    \item MLP-based models: DLinear ~\cite{zengAreTransformersEffective2022}, NLinear ~\cite{zengAreTransformersEffective2022}, LightTS ~\cite{zhangLessMoreFast2022}.
    \item CNN-based models: TimesNet ~\cite{wuTimesNetTemporal2DVariation2023}.
\end{itemize}
% Random Seed

\textbf{Other Settings} Our experiments are conducted on a NVIDIA GeForce GTX 1080 Ti. Moreover, experiments are implemented in PyTorch ~\cite{PyTorchImperativeStyle2019paszke} 1.10.1. Our method is trained with Huber Loss~\cite{huber1964robust}, using the Adam optimizer ~\cite{AdamMethodStochastic2015kingma}.
%模型参数的设置，学习率, Early stop, conv参数初始化方式

\subsection{Main Results}
\begin{table*}[htbp]
  \centering
  \caption{Univariate long-term forecasting on the ETT datasets.}
  \resizebox{0.9\textwidth}{!}{
    \begin{tabular}{c|c|cc|cc|cc|cc|cc|cc|cc|cc}
    \toprule
    \multicolumn{2}{c|}{Models}& \multicolumn{2}{c|}{MSDCN}& \multicolumn{2}{c|}{TimesNet}& \multicolumn{2}{c|}{NLinear}& \multicolumn{2}{c|}{DLinear}& \multicolumn{2}{c|}{LightTS}& \multicolumn{2}{c|}{ETSformer}& \multicolumn{2}{c|}{Autoformer}& \multicolumn{2}{c}{Informer} \\
    \midrule
    \multicolumn{2}{c|}{Metric}& MSE& MAE& MSE& MAE& MSE& MAE& MSE& MAE& MSE& MAE& MSE& MAE& MSE& MAE& MSE& MAE \\
    \midrule
    \multirow{4}[2]{*}{\rotatebox{90}{ETTh1}}& 96& \textbf{0.052}& \textbf{0.177}& 0.059& 0.189& 0.055& 0.179& 0.056& 0.183& \underline{0.053}& \underline{0.177}& 0.056& 0.180& 0.071& 0.206& 0.193& 0.377  \\
      & 192& \textbf{0.067}& \textbf{0.202}& 0.074& 0.215& 0.071& 0.205& 0.072& 0.208& \underline{0.069}& \underline{0.204}& 0.071& 0.204& 0.114& 0.262& 0.217& 0.395  \\
      & 336& \underline{0.078}& \underline{0.225}& \textbf{0.076}& \textbf{0.220}& 0.081& 0.225& 0.083& 0.225& 0.081& 0.226& 0.098& 0.244& 0.107& 0.258& 0.202& 0.381  \\
      & 720& \textbf{0.077}& \textbf{0.220}& 0.087& 0.236& 0.087& 0.232& 0.089& 0.236& \underline{0.080}& \underline{0.226}& 0.189& 0.359& 0.126& 0.283& 0.183& 0.355  \\ % \cmidrule{2-18}
     % & Avg& \textbf{0.068}& \textbf{0.206}& 0.074& 0.215& 0.074& 0.210& 0.075& 0.213& \underline{0.071}& \underline{0.208}& 0.104& 0.247& 0.105& 0.252& 0.199& 0.377  \\
    \midrule
    \multirow{5}[2]{*}{\rotatebox{90}{ETTh2}}& 96& \textbf{0.118}& \textbf{0.269}& 0.131& 0.284& 0.129& 0.282& 0.136& 0.286& \underline{0.129}& \underline{0.278}& 0.131& 0.279& 0.153& 0.306& 0.213& 0.373  \\
      & 192& \textbf{0.155}& \textbf{0.312}& 0.171& 0.329& \underline{0.168}& 0.328& 0.182& 0.336& 0.169& \underline{0.324}& 0.176& 0.329& 0.204& 0.351& 0.227& 0.387  \\
      & 336& \underline{0.174}& \textbf{0.335}& \textbf{0.171}& \underline{0.336}& 0.185& 0.351& 0.216& 0.369& 0.194& 0.355& 0.209& 0.367& 0.246& 0.389& 0.242& 0.401  \\
      & 720& \textbf{0.196}& \textbf{0.355}& \underline{0.223}& \underline{0.380}& 0.224& 0.383& 0.245& 0.396& 0.225& 0.381& 0.276& 0.426& 0.268& 0.409& 0.291& 0.439  \\  %\cmidrule{2-18}
      %& Avg& \textbf{0.161}& \textbf{0.318}& \underline{0.174}& \underline{0.332}& 0.177& 0.336& 0.195& 0.347& 0.179& 0.335& 0.198& 0.350& 0.218& 0.364& 0.243& 0.400  \\
    \midrule
    \multirow{5}[2]{*}{\rotatebox{90}{ETTm1}}& 96& \textbf{0.026}& \underline{0.122}& \underline{0.026}& 0.123& 0.026& \textbf{0.121}& 0.029& 0.127& 0.026& 0.122& 0.028& 0.123& 0.056& 0.183& 0.109& 0.277  \\
      & 192& \underline{0.040}& 0.151& 0.040& 0.151& \textbf{0.039}& \underline{0.150}& 0.046& 0.162& 0.039& \textbf{0.149}& 0.045& 0.156& 0.081& 0.216& 0.151& 0.310  \\
      & 336& \textbf{0.052}& 0.175& 0.053& 0.174& 0.053& \underline{0.173}& 0.060& 0.188& 0.052& \textbf{0.172}& 0.061& 0.182& 0.076& 0.218& 0.427& 0.591  \\
      & 720& \textbf{0.072}& 0.208& \underline{0.073}& \textbf{0.206}& 0.074& \underline{0.207}& 0.081& 0.219& 0.073& 0.207& 0.080& 0.210& 0.110& 0.267& 0.438& 0.586  \\ % \cmidrule{2-18}
     % & Avg& \textbf{0.047}& 0.164& \underline{0.048}& 0.164& 0.048& \textbf{0.163}& 0.054& 0.174& 0.048& \underline{0.163}& 0.054& 0.168& 0.081& 0.221& 0.281& 0.441  \\
    \midrule
    \multirow{4}[2]{*}{\rotatebox{90}{ETTm2}}& 96& \textbf{0.062}& \textbf{0.182}& 0.065& 0.187& 0.065& 0.186& 0.066& 0.186& \underline{0.063}& \underline{0.182}& 0.063& 0.183& 0.065& 0.189& 0.088& 0.225  \\
      & 192& \underline{0.091}& \underline{0.226}& 0.093& 0.231& 0.094& 0.231& 0.102& 0.240& \textbf{0.090}& \textbf{0.223}& 0.092& 0.227& 0.118& 0.256& 0.132& 0.283  \\
      & 336& \textbf{0.091}& \textbf{0.226}& 0.121& 0.266& 0.120& 0.265& 0.132& 0.277& \underline{0.117}& \underline{0.259}& 0.119& 0.261& 0.154& 0.305& 0.180& 0.336  \\
      & 720& \textbf{0.166}& \underline{0.319}& 0.172& 0.322& 0.171& 0.322& 0.185& 0.336& \underline{0.170}& \textbf{0.318}& 0.175& 0.320& 0.182& 0.335& 0.300& 0.435  \\  %\cmidrule{2-18}
      %& Avg& \textbf{0.103}& \textbf{0.238}& 0.113& 0.252& 0.113& 0.251& 0.121& 0.260& \underline{0.110}& \underline{0.246}& 0.112& 0.248& 0.130& 0.271& 0.175& 0.320  \\
    \bottomrule
    \end{tabular}%
    }
  \label{tab:main2}%
%}
\end{table*}%

% 写明实验数据的结果是从哪拿过来的，或者是自己运行代码得到的, 在正文写
% 展示实验结果；
To show the effectiveness of our model on benchmark datasets, we conduct both multivariate and univariate long-term forecasting. The results are summarized as follows.
% 并且我们发现仅用我们的方法仅用一层卷积就可以比TimesNet的两层堆叠卷积效果更好
%为了更清晰地展示结果，我们将ETT数据集的结果和其他领域的数据集分开展示。
\subsubsection{Multivariate Long-term Forecasting}
In order to present the results more clearly, we present the results of in Table \ref{tab:main1}, and full benchmark on ETT datasets are showned in supplementary materials. 
%On the ETT dataset, our approach demonstrates significant improvements compared to the best-performing NLinear model and the second-best TimesNet model. 
%Compared to the ETT dataset, Weather, ECL, and Traffic datasets have a larger number of variables, and illness dataset exhibits greater volatility. 
%On those datasets, our method shows slight improvements over the best model, TimesNet. 
%This might be attributed to the fact that our model has a smaller model capacity compared to TimesNet, limiting the extent of potential improvement.
%Taking into account the improvements based on the statistical results from both tables, we can observe that, first,
Our model achieves the best results in 80\%  of prediction results.
%Second, our model performs effectively across datasets from diverse domains.
Compared to the best CNN-based model, TimesNet, our model exhibits a reduction of 2.4\% in the MSE metric and a reduction of 2.2\% in the MAE metric. These results indicate that 
our design of shallow multi-scale 1D dilated convolutional modules outperforms deeper 2D ordinary convolution. Our method uses dilated convolution to downsample time series data at different scales, which can better capture multi-scale information.

When compared to the best transformer-based model, ETSformer, our model demonstrates a decrease of 9.4\% in the MSE metric and a decrease of 11.0\% in the MAE metric. 
After the TimesNet model, our model once again demonstrates that in long-term time series forecasting, the use of  convolution networks can be more effective than transformer architectures.
Furthermore, when compared to the best MLP-based model, NLinear, our model showcases a decrease of 9.4\% in the MSE metric and a decrease of 4.2\% in the MAE metric. 
The NLinear model adopts an autoregressive structure with simple normalization operations before input and after output. 
Our model can be viewed as an extension of NLinear, incorporating our specifically designed convolutional modules. 
The results indicate that those convolutional modules enhance the autoregressive module expressive capabilities. 

\subsubsection{Univariate Long-term Forecasting}  In the convolutional modules, we utilize channel independence to model each variable separately.
However, in the feed forward layer, multiple variables share the same model parameters. 
To validate the effectiveness of our model in single-variable prediction, We select the Oil Temperature column in the ETT datasets as the univariate target. and the results are shown in Table \ref{tab:main2}. Following past studies, we set the lookback window to 336. Compared to the NLinear model, our proposed method reduces the MSE by 4.8\%. Against the PatchTST/64 model, it decreases by 6.8\%, and compared to the TimesNet model, it drops by 12.9\%. These results shows that MSDCN achieves superior performance on univariate long  sequences forecasting.

\subsection{Ablation Study}
%\subsubsection{Convolutional modules}
\subsubsection{Convolutional Modules} 
\begin{table}[tbp]
  \centering 
  \caption{Multivariate long-term forecasting with or without convolutional module.}
  \label{fig:conv}
  \setlength{\tabcolsep}{0.5mm}{
  \resizebox{\linewidth}{!}{
 % \resizebox{0.9\textwidth}{!}{
 % \resizebox{0.85\textwidth}{!}{
       \begin{tabular}{p{2.6em}|p{2.6em}|c|cccc|c|cccc|c}
    \toprule
    \multicolumn{2}{p{5.4em}|}{Convolution} & \multirow{3}[4]{*}{Metric} & \multicolumn{5}{c|}{\multirow{2}[2]{*}{ETTm2}} & \multicolumn{5}{c}{\multirow{2}[2]{*}{ECL}} \\
    \multicolumn{2}{p{5.4em}|}{\ \ Module} &     & \multicolumn{5}{c|}{}       & \multicolumn{5}{c}{} \\
\cmidrule{1-2}\cmidrule{4-13}   \  long &\ short &     & 96  & 192 & 336 & 720 & Avg & 96  & 192 & 336 & 720 & Avg \\
    \midrule
    \multirow{2}[2]{*}{\ \ $\times$} & \multirow{2}[2]{*}{\ \ $\times$} & MSE & 0.182  & 0.246  & 0.306  & 0.408  & 0.286  & 0.198  & 0.198  & 0.212  & 0.254  & 0.216  \\
        &     & MAE & 0.265  & 0.304  & 0.341  & 0.398  & 0.327  & 0.275  & 0.278  & 0.293  & 0.326  & 0.293  \\
    \midrule
    \multirow{2}[2]{*}{\ \ $\checkmark$} & \multirow{2}[2]{*}{\ \ $\times$} & MSE & 0.176  & 0.239  & 0.300  & 0.398  & 0.278  & 0.176  & 0.184  & 0.200  & 0.241  & 0.200  \\
        &     & MAE & 0.257  & 0.299  & 0.335  & \textbf{0.392 } & 0.321  & 0.265  & 0.272  & 0.287  & 0.321  & 0.286  \\
    \midrule
    \multirow{2}[2]{*}{\ \ $\times$} & \multirow{2}[2]{*}{\ \ $\checkmark$} & MSE & 0.177  & 0.242  & 0.302  & 0.407  & 0.282  & 0.194  & 0.193  & 0.208  & 0.250  & 0.211  \\
        &     & MAE & 0.256  & 0.300  & 0.336  & 0.399  & 0.323  & 0.271  & 0.274  & 0.289  & 0.322  & 0.289  \\
    \midrule
    \multirow{2}[2]{*}{\ \ $\checkmark$} & \multirow{2}[2]{*}{\ \ $\checkmark$} & MSE & \textbf{0.174} & \textbf{0.239} & \textbf{0.296} & \textbf{0.395} & \textbf{0.276} & \textbf{0.175} & \textbf{0.183} & \textbf{0.199} & \textbf{0.239} & \textbf{0.199} \\
        &     & MAE & \textbf{0.256} & \textbf{0.298} & \textbf{0.332} & 0.392  & \textbf{0.320} & \textbf{0.265} & \textbf{0.271} & \textbf{0.287} & \textbf{0.320} & \textbf{0.286} \\
    \bottomrule
    \end{tabular}%
   }
   }
  %  }
\end{table}%
We conduct experiments to analyze the effects of convolutional modules, as shown in Table \ref{fig:conv}. Without long and short convolutional modules denotes that we remove the entire convolutional modules and keep only the autoregressive module.
% ht or bp

%By comparing the results with and without long or short convolutional module, we observed that both the long and short convolutional modules have a positive effect. 
%The long convolutional module exhibits superior performance when compared to solely adding the short convolutional module. 
%Furthermore, the combination of both modules result in even better performance.
The table demonstrates that incorporating both long and short convolution modules enhances prediction outcomes. 
For the ETTm2 dataset with a 15-minute sampling interval, the impact on predictions is less noticeable due to the dataset's sensitivity to short-term factors. 
Conversely, for the Electricity dataset with a 1-hour sampling interval, solely adding the long convolution module yields notably better predictions than solely adding the short convolution module. 
This is attributed to the Electricity dataset being significantly influenced by long-term factors. 
In conclusion, introducing both modules simultaneously achieves optimal prediction performance.

\subsubsection{Autoregressive}
\begin{figure}[tbp]
	\centering
	\begin{subfigure}{0.49\linewidth}
		\centering
		\includegraphics[width=\linewidth]{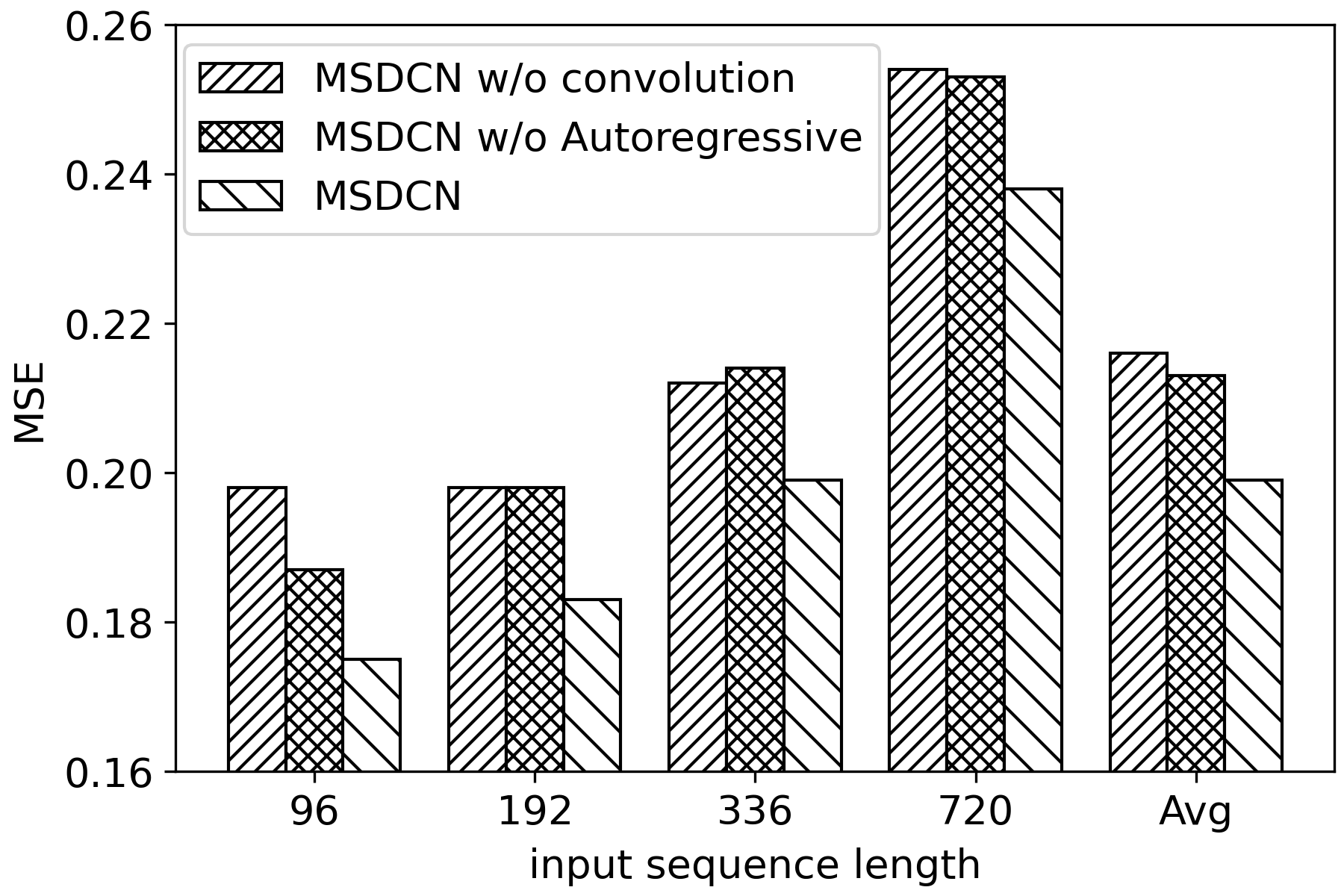}
          \caption{MSE on ECL dataset.}
	\end{subfigure}
 \begin{subfigure}{0.49\linewidth}
		\centering
		\includegraphics[width=\linewidth]{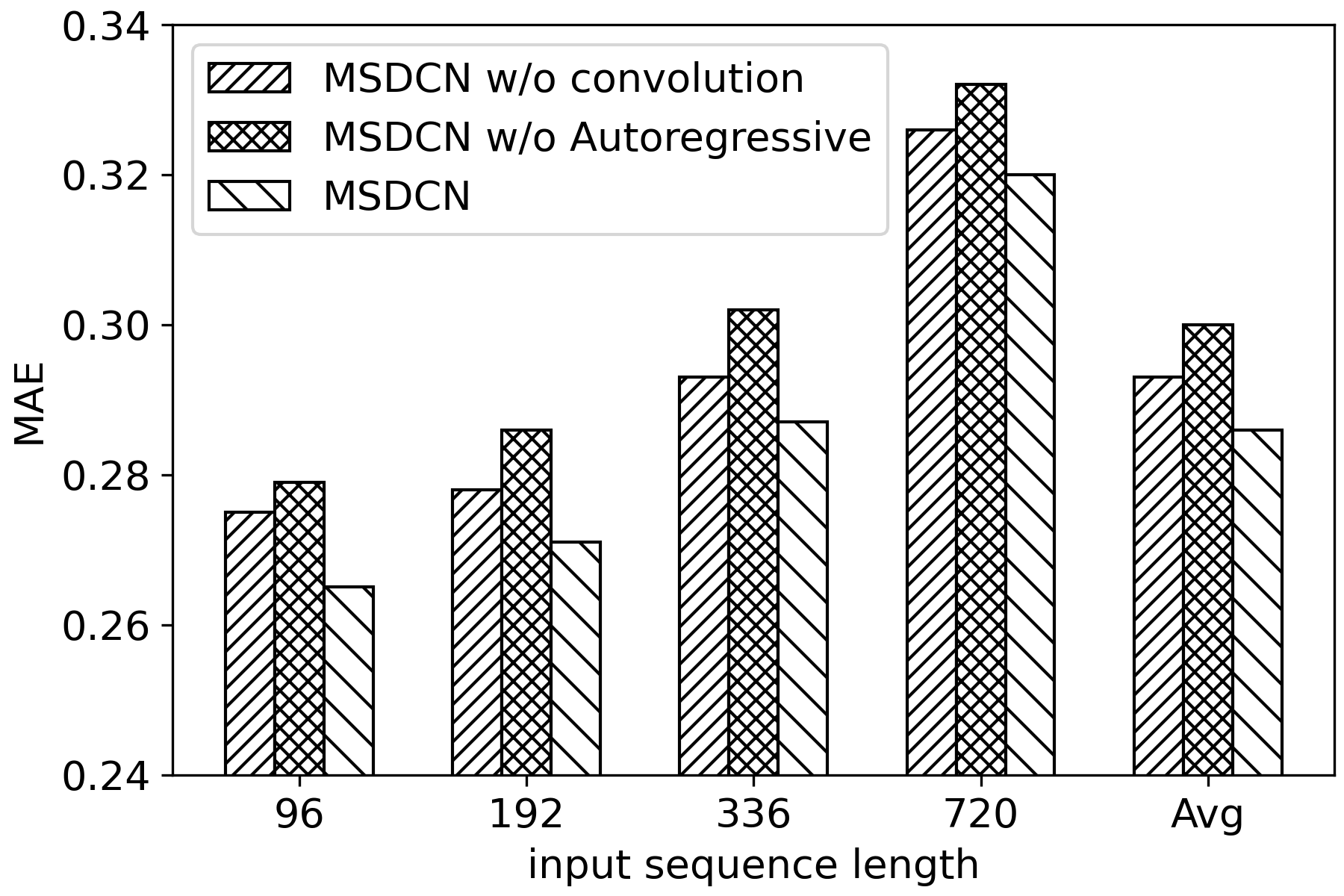}
          \caption{MAE on ECL dataset.}
	\end{subfigure}
 \begin{subfigure}{0.49\linewidth}
		\centering
		\includegraphics[width=\linewidth]{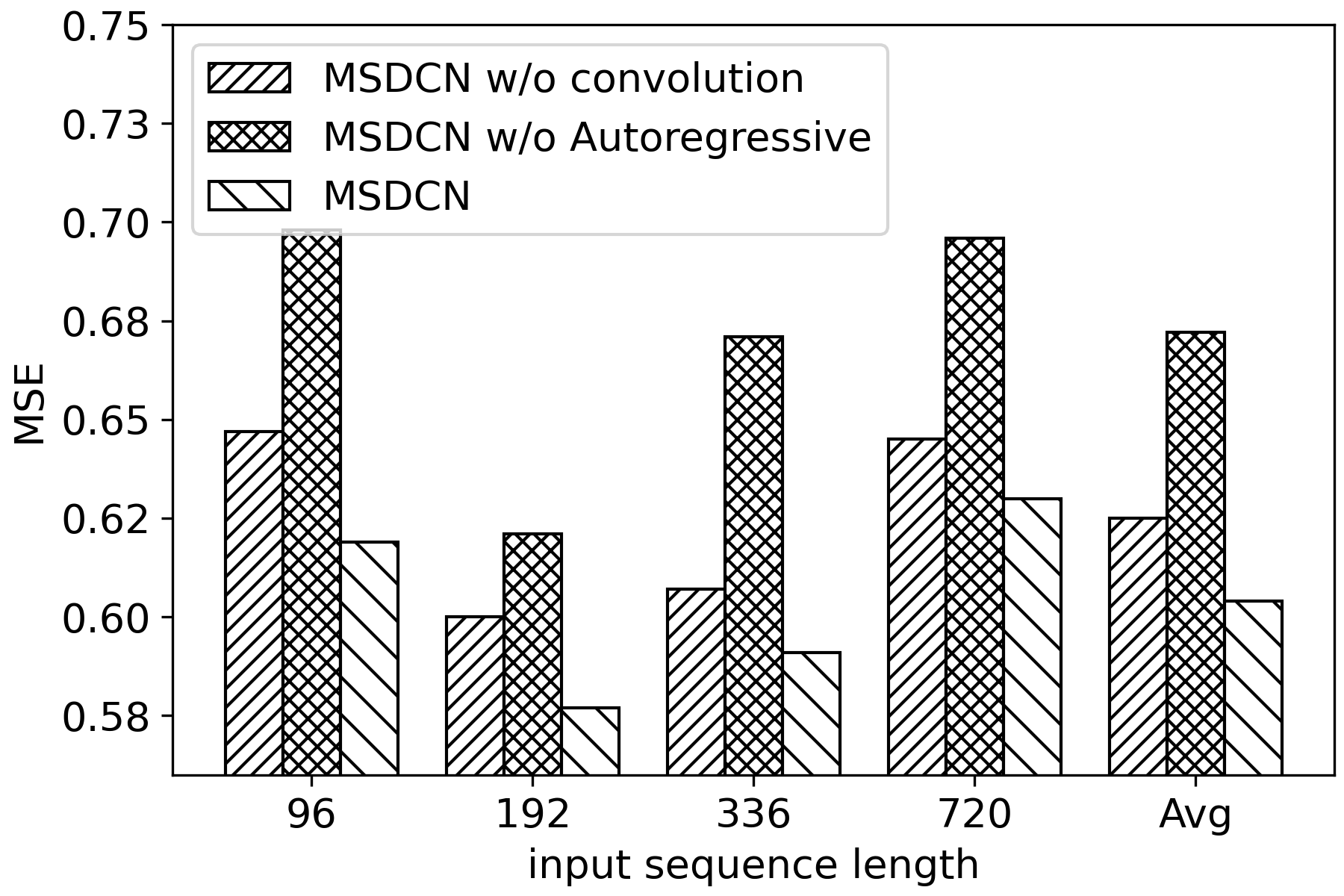}
  \caption{MSE on traffic dataset.}
	\end{subfigure}
 \begin{subfigure}{0.49\linewidth}
		\centering
		\includegraphics[width=\linewidth]{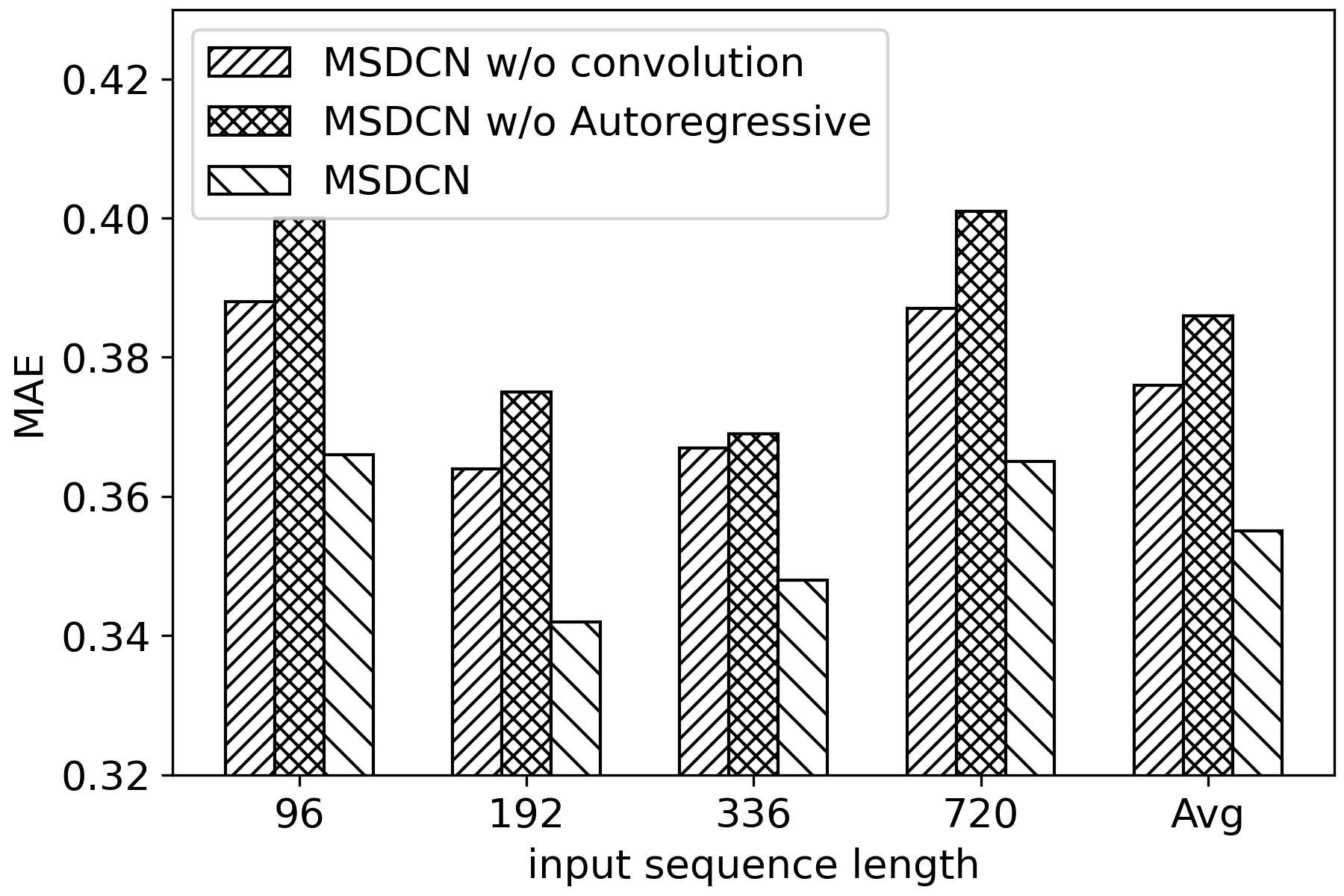}
  \caption{MAE on traffic dataset.}
	\end{subfigure}
	\caption{Ablation results for the autoregressive module in MSDCN.}
	\label{fig:autogressive}
\end{figure}
%\textbf{Autoregressive.} 
To validate the impact of autoregressive module, we compare different architectures: MSDCN without convolution modules (only autoregressive module), MSDCN without autoregressive module (only convolution module), and the complete MSDCN, which integrates both modules. 
The results are shown in Figure \ref{fig:autogressive}. 

The results show that using only the autoregressive module gives reasonable predictions. 
However, when we add the convolution module, the overall network performance improves. 
Specifically, we see a 7\% improvement in the Electricity dataset and a 3\% enhancement in the Traffic dataset. 
This highlights the autoregressive module's ability to catch linear relationship, making predictions more accurate. 
The autoregressive is only a linear layer and cannot deal with complex temporal patterns. The introduced convolution module in MSDCN tackles this issue by capturing intricate patterns and trends in time series data, leading to better forecast accuracy. This confirms the effectiveness of our proposed method.
% 再加一个表对比，加上autoregressive的方法和我们的不同
% 对比方法LSTNet, NLinear, LightTS, 产生反作用，我们的可以起到正向作用

\section{Analysis} \label{5}
% 主要注重TimesNet、PatchTST方法与我们方法的对比
\subsection{Multi-Scale Representation}
\begin{figure}[tbp]
	\centering
	\begin{subfigure}{0.495\linewidth}
		\centering
		\includegraphics[width=\linewidth]{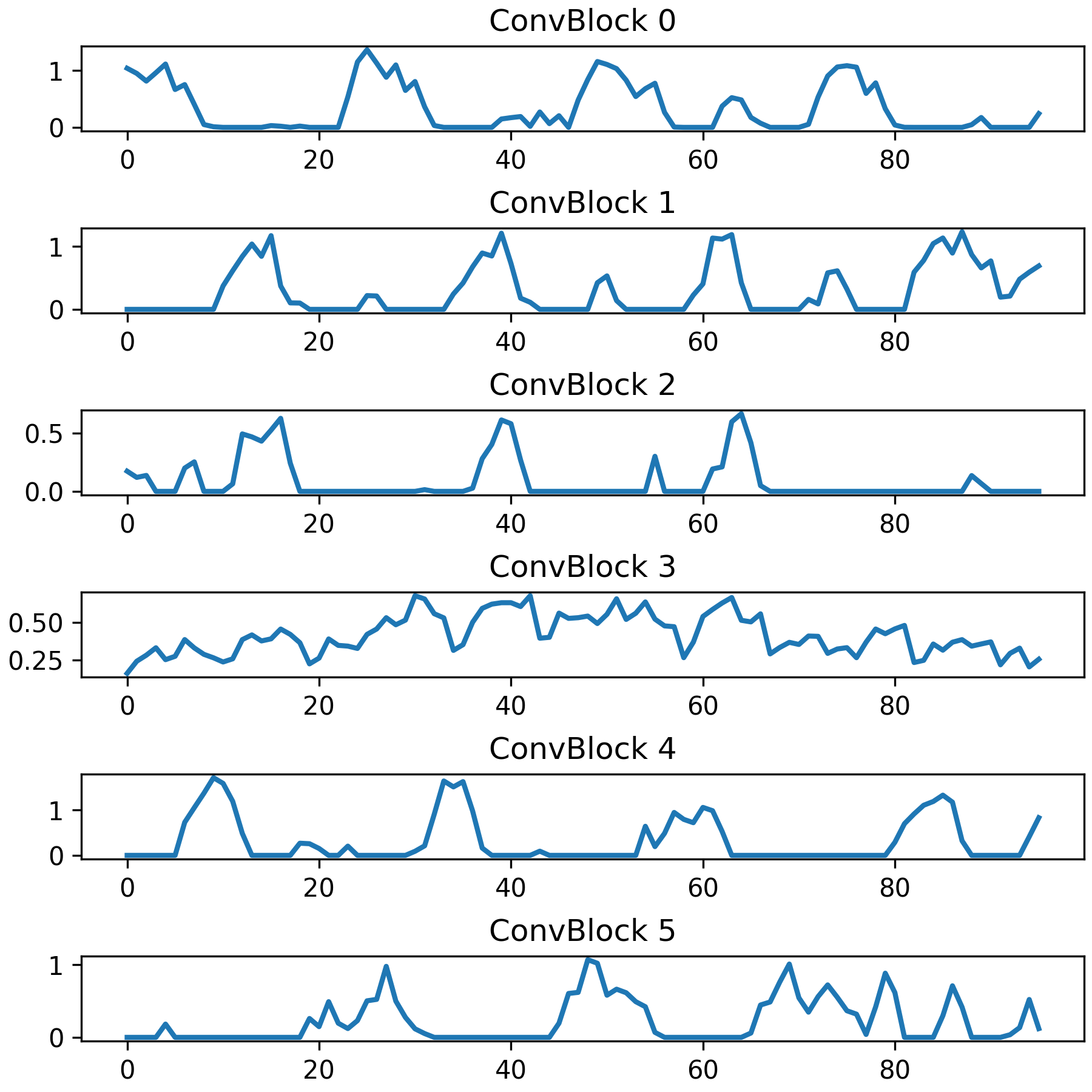}
      \caption{ECL case}
	\end{subfigure}
  \begin{subfigure}{0.495\linewidth}
		\centering
		\includegraphics[width=\linewidth]{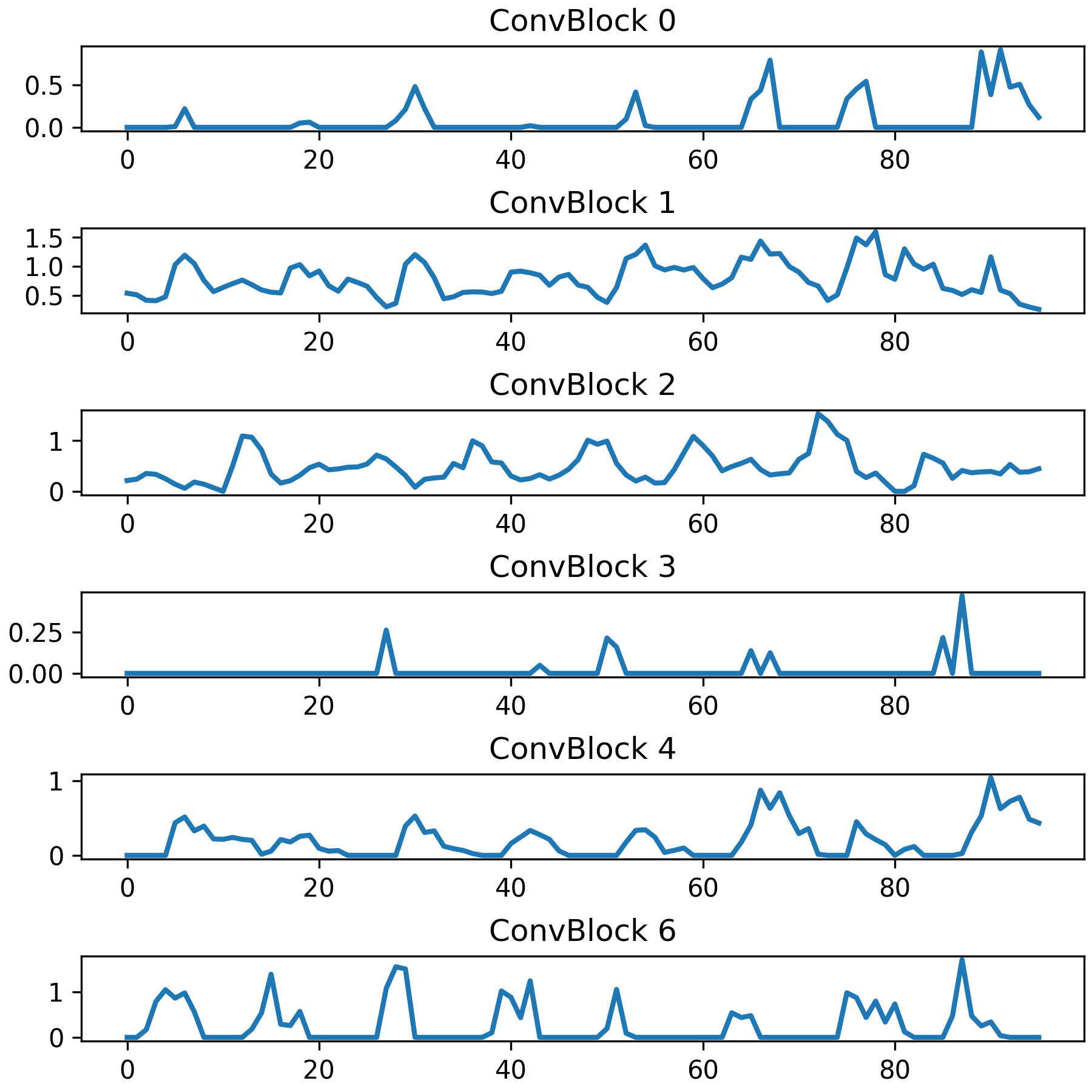}
   \caption{traffic case}
	\end{subfigure}
	\caption{Visualization of different convolutional block output representation. The prediction length is 336. }
	\label{fig:conv_learn}
\end{figure}

We visualize the outputs from various convolution blocks, as shown in Figure \ref{fig:conv_learn}.
%For the ECL dataset, the network consists of 2 blocks in the short convolutional module and 4 blocks in the long convolutional module. For the traffic dataset, we use 3 blocks in the short convolutional module and use 4 blocks in the long convolutional module. 
The horizontal axis in the figure represents the time steps of the input time series, while the vertical axis represents the values after convolution. 
Each curve in the subfigures represents a different scale of convolution block representation, arranged from top to bottom. 
We use ReLU function in convolution blocks, so some negative values are suppressed. 
Blocks with missing IDs indicate that all values within those blocks are suppressed.

From Figure \ref{fig:conv_learn}, we can observe that the different scales of convolution blocks can extract various curves with different shapes, indicating that the periods and trends of these curves are distinct, which demonstrates the effectiveness of our method. 
Therefore, our method exhibits effectiveness and robustness in handling diverse features of different scales in time series data.

 \begin{figure}[tbp]
	\centering
	\begin{subfigure}{0.495\linewidth}
		\centering
		\includegraphics[width=\linewidth]{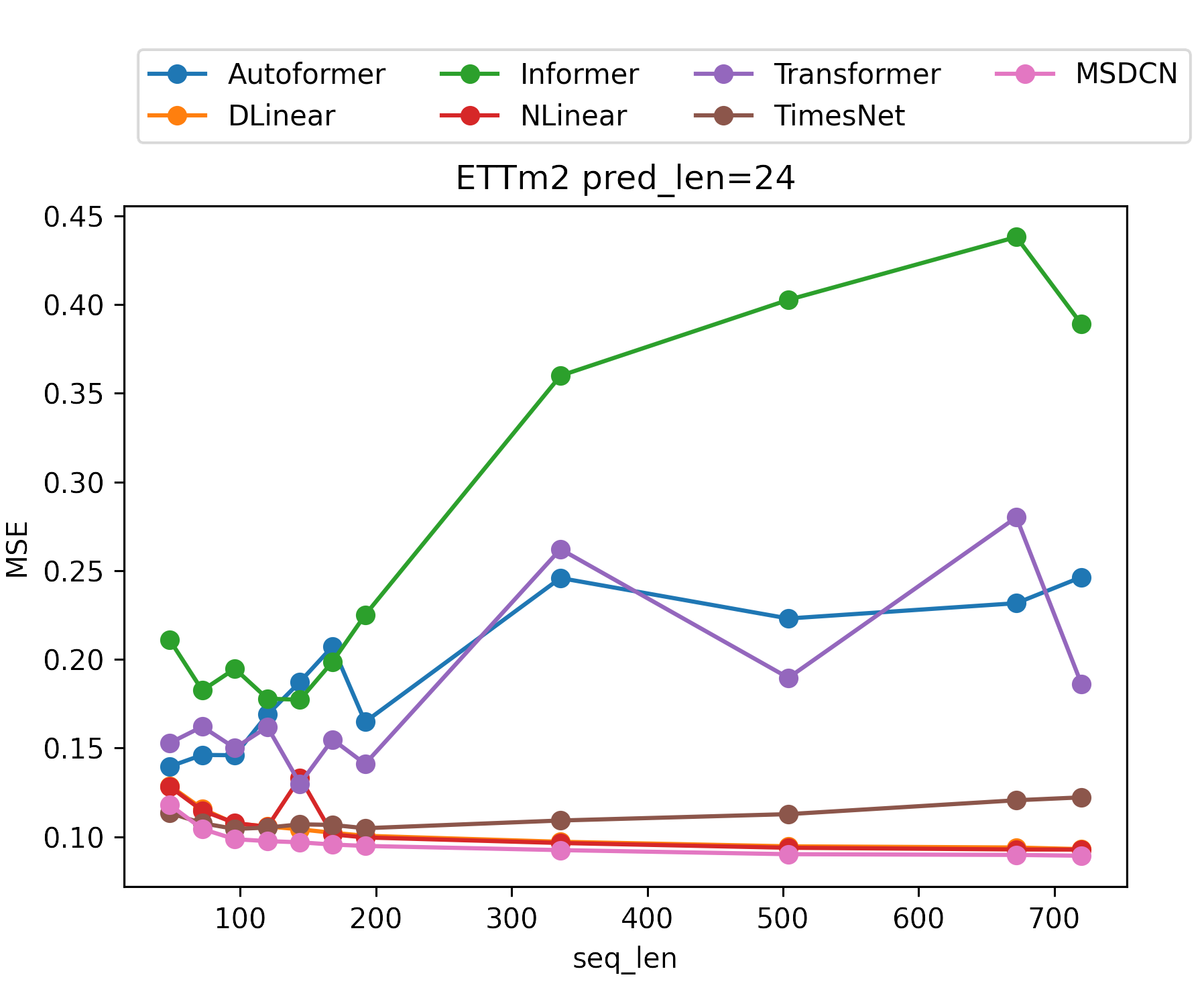}
	\end{subfigure}
 \begin{subfigure}{0.495\linewidth}
		\centering
		\includegraphics[width=\linewidth]{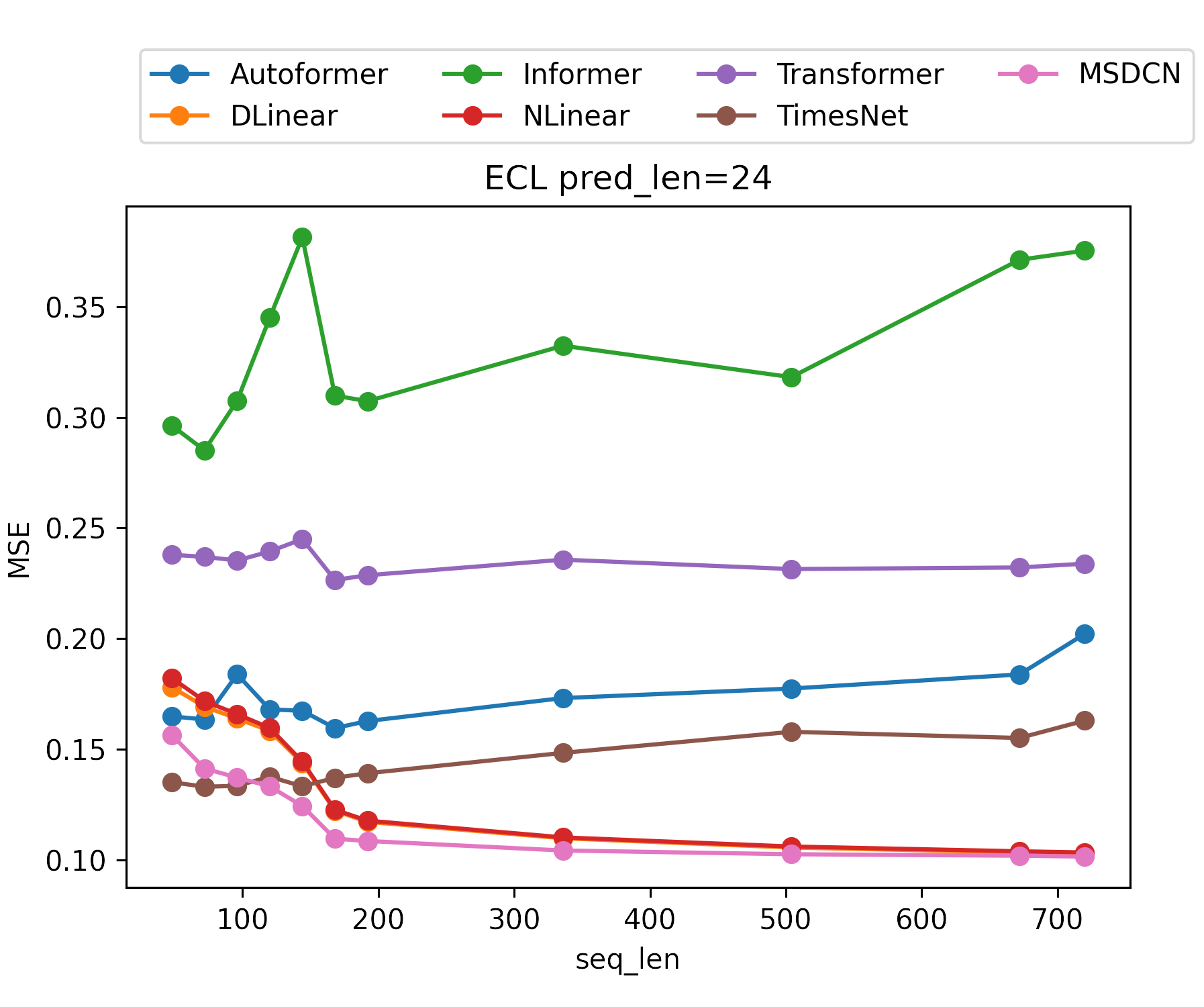}
	\end{subfigure}
  \begin{subfigure}{0.495\linewidth}
		\centering
		\includegraphics[width=\linewidth]{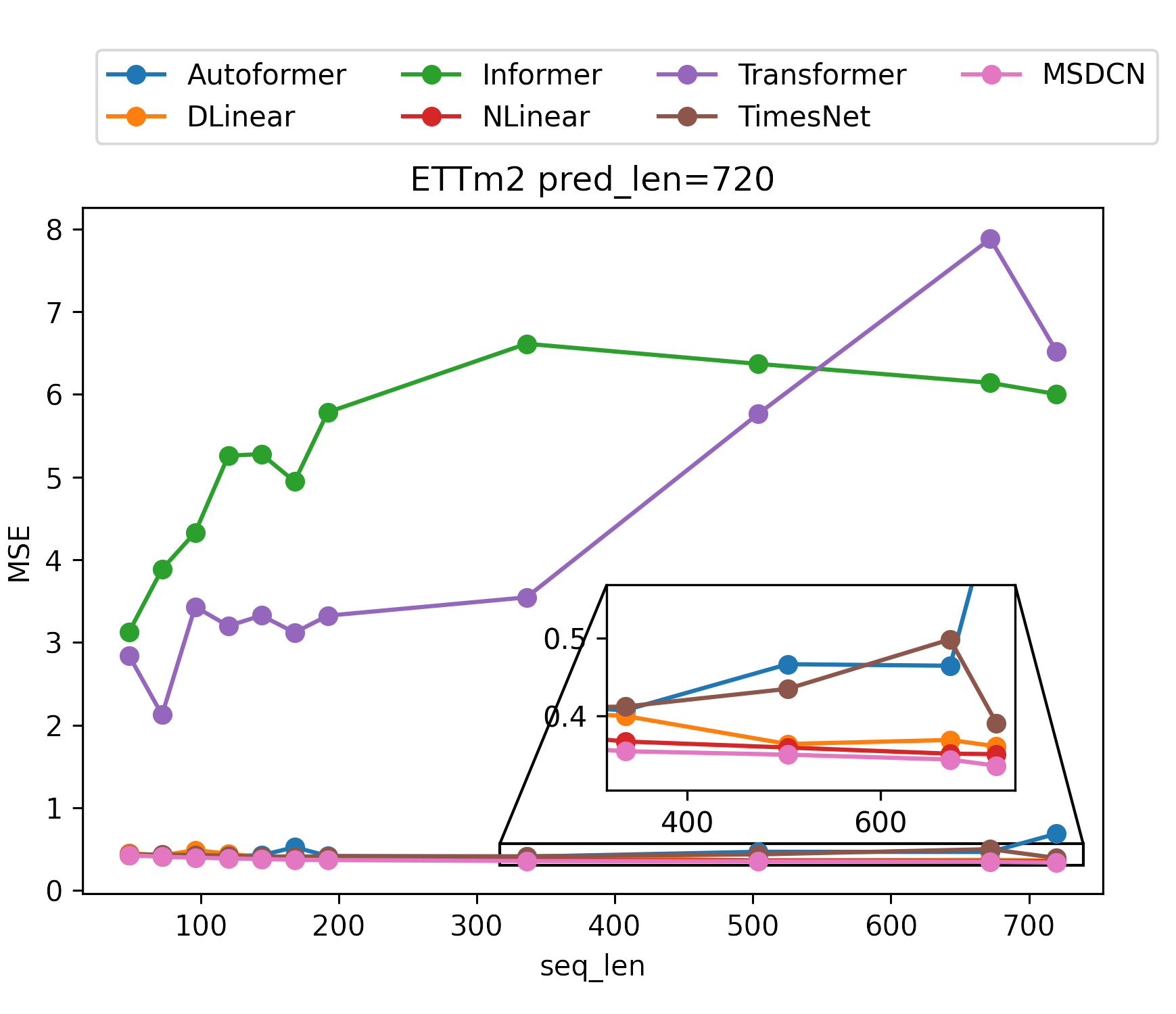}
	\end{subfigure}
 \begin{subfigure}{0.495\linewidth}
		\centering
		\includegraphics[width=\linewidth]{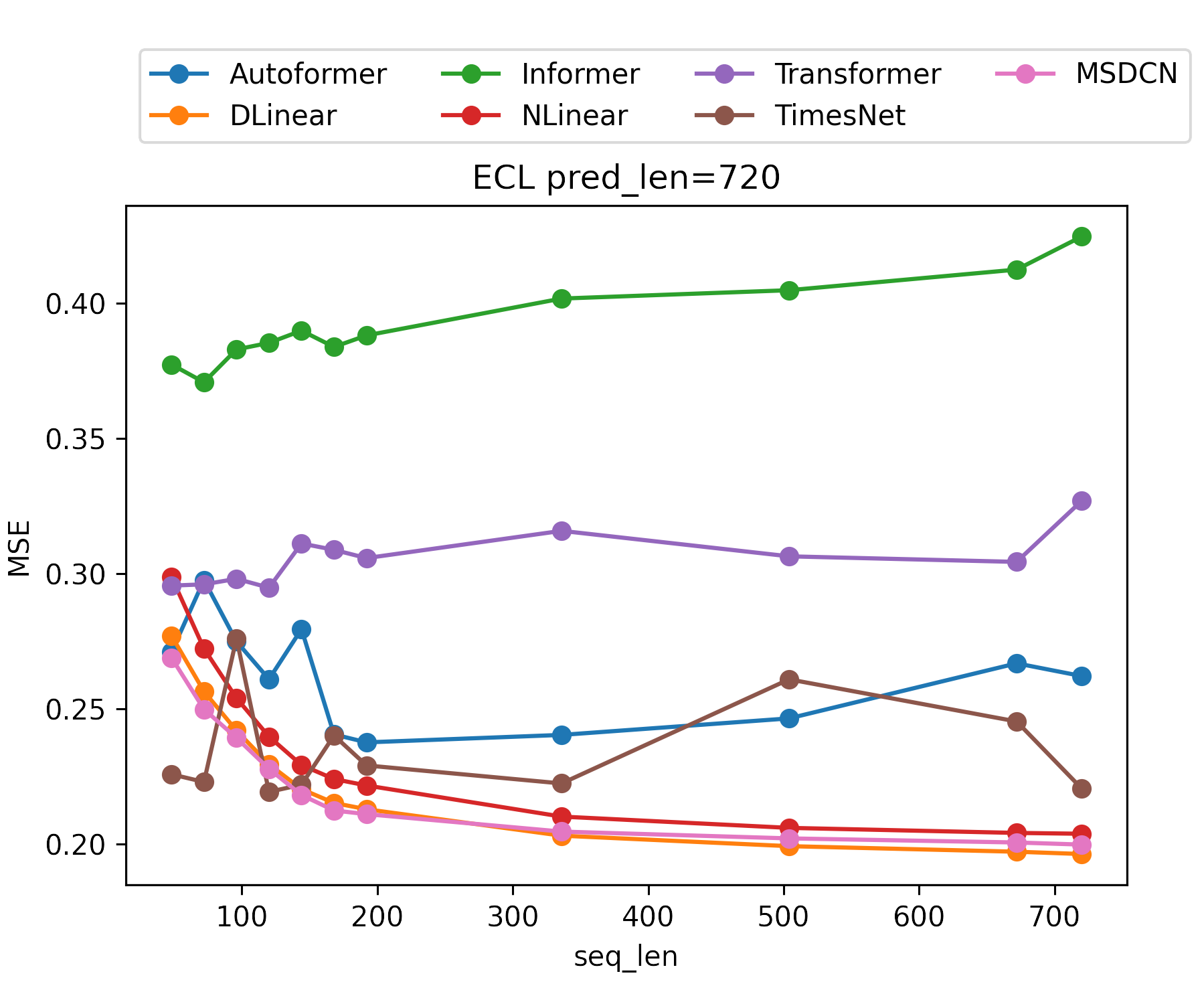}
	\end{subfigure}
	\caption{The MSE performance when input length is increasing. The prediction sequence length is 24 or 720. }
	\label{fig:diff_input_len}
\end{figure}
%卷积特征图可视化
% 输入长度的增加，预测性能对比
\subsection{Different Input Lengths On Prediction Performance}

% 理想情况下，在固定的输出长度下，随着输入序列长度的增加，模型可以看到更多的信息，提取到更多的时间特性，能够做更有效地预测。一个好的模型是随着输入序列长度的增加，预测效果不会下降。如图\ref{fig:diff_input_len}所示，我们对比了几个比较典型的模型与我们提出的模型，在输出长度为24或720，输入序列长度为{48, 72, 96, 120, 144, 168, 192, 336, 504, 672, 720}下的MSE表现。在输入序列长度增加后，TimesNet等模型的性能出现了下降，出现了过拟合现象，更多地学习到了数据中的噪声，无法从较长的输入中提取出有效的时序模式。此外，我们提出的模型相比于DLinear, NLinear等模型，在输入序列长度较短时，效果更好一点，并且输入序列长度较长时，也不会变差。
With a fixed output length, as the length of the input sequence increases, the model can access more information, extract additional temporal features, and make more effective predictions. 
A good model should maintain consistent prediction performance without a decline as the input sequence length increases.

As shown in Figure \ref{fig:diff_input_len}, 
%We conduct experiments on several typical models with an output length of 24 or 720 and input sequence lengths of {48, 72, 96, 120, 144, 168, 192, 336, 504, 672, 720}. 
as the input sequence length increases, models like TimesNet exhibit a decline in performance and suffer from overfitting, as they tend to learn more noise from the data rather than extracting meaningful temporal patterns from longer inputs. 
Our method outperforms models like DLinear when the input sequence length is relatively short and maintains its performance even with longer input sequences. 
Moreover, DLinear and NLinear demonstrate remarkably similar performance under most conditions. 
Our model exhibits superior performance when the input sequence is relatively short. 
As the length of input sequence increases, our model outperforms DLinear and NLinear by a smaller margin.

\subsection{Efficiency Analysis}
\begin{table}[tbp]
  \centering
  \caption{Comparison of practical efficiency of Models under input\_len=96 and output\_len=720 on the Traffic. Memory is one forward/backward pass size. MACs are the number of multiply-accumulate operations. Time is the inference time.}
  \resizebox{\linewidth}{!}{
  %\resizebox{0.8\textwidth}{!}{
    \begin{tabular}{c|c|c|c|c}
    \toprule
    Model & Total Params & Memory & MACs & Time \\
    \midrule
    MSDCN & \underline{239.67K} & \underline{19.16MB} & \underline{8.75M} & \underline{3.68s} \\
    \midrule
    DLinear & \textbf{139.68K} & \textbf{13.48MB} & \textbf{0.14M} & \textbf{1.50s} \\
    \midrule
    Autoformer & 14.91M & 126.33MB & 4.18G & 62.39s \\
    \midrule
    PatchTST & 7.59M & 800.96MB & 12.69G & 80.42s \\
    \midrule
    TimesNet & 301.75M & 1628.31MB & 1.23T & 1491.58s \\
    \bottomrule
    \end{tabular}%
    }
  \label{tab:complex}%
\end{table}%
%模型参数量比较；模型计算量比较；训练时间；推理时间
%分析各个模型为什么推理时间长，计算量大，参数量大
% 我们比较了目前较好模型的实际效率。我们的模型仅次于最简单的结构DLinear模型，比基于Transformer的结构要快。更重要的是，比同样基于CNN的模型——TimesNet，推理速度快了很多。
We compare our model with benchmark models based on the average practical efficiencies over 5 runs, as shown in Table \ref{tab:complex}. 
DLinear has only two linear layers, making it the fastest in terms of speed. Our model is second only to DLinear and is faster than models based on Transformer architecture. 

More importantly, our model has significantly improved the inference speed compared to the CNN-based TimesNet model. TimesNet first transforms the data through an embedding layer and obtain representations through parameter-efficient Inception blocks, resulting in a significant increase in its network parameters and computational complexity. 
Compared with TimesNet, our model does not have an embedding layer and use shallow convolution layer to extract representation, therefore, our model's inference time is faster than TimesNet model.

\section{Conclusions}\label{6}
This paper presents a novel CNN-based neural network on long-term time series forecasting. 
To extract complex temporal dependencies in long-term time series, MSDCN can extract multi-scale temporal features by using different dilated convolution blocks. Extensive experiments showcase the effectiveness of MSDCN on prediction accuracy and time efficiency.

%In future work, we will explore the question that how to judge whether variables have an implicit relationship and how to model the spatial relationship between auxiliary and target variables in long-term time series forecasting.

%\section{Acknowledgments}
%This work was supported in part by the STI 2030-Major Projects of China under Grant 2021ZD0201300, and by the National Science Foundation of China under Grant 62276127.


%\bibliographystyle{elsarticle-num} 


\bibliographystyle{named}
\bibliography{ijcai24}

\newpage

\appendix

\end{document}